\documentclass[acmsmall]{acmart}
\AtBeginDocument{%
  \providecommand\BibTeX{{%
    \normalfont B\kern-0.5em{\scshape i\kern-0.25em b}\kern-0.8em\TeX}}}

\setcopyright{acmcopyright}
\copyrightyear{2023}
\acmYear{2023}
\acmDOI{XXXXXXX.XXXXXXX}

\acmJournal{JACM}
\acmVolume{37}
\acmNumber{4}
\acmArticle{111}
\acmMonth{8}

\begin{document}

%%
%% The "title" command has an optional parameter,
%% allowing the author to define a "short title" to be used in page headers.
\title{A Survey on Recent Teacher-student Learning Studies}

%%
%% The "author" command and its associated commands are used to define
%% the authors and their affiliations.
%% Of note is the shared affiliation of the first two authors, and the
%% "authornote" and "authornotemark" commands
%% used to denote shared contribution to the research.
\author{Gao Minghong}
\email{1374278204@qq.com}
\affiliation{%
  \institution{Northwestern Polytechnic University}
  \country{China}
}

%%
%% By default, the full list of authors will be used in the page
%% headers. Often, this list is too long, and will overlap
%% other information printed in the page headers. This command allows
%% the author to define a more concise list
%% of authors' names for this purpose.
\renewcommand{\shortauthors}{Gao, et al.}

%%
%% The abstract is a short summary of the work to be presented in the
%% article.
\begin{abstract}
Knowledge distillation is a method of transferring the knowledge from a complex deep neural network (DNN) to a smaller and faster DNN, while preserving its accuracy. Recent variants of knowledge distillation include teaching assistant distillation, curriculum distillation, mask distillation, and decoupling distillation, which aim to improve the performance of knowledge distillation by introducing additional components or by changing the learning process. Teaching assistant distillation involves an intermediate model called the teaching assistant, while curriculum distillation follows a curriculum similar to human education. Mask distillation focuses on transferring the attention mechanism learned by the teacher, and decoupling distillation decouples the distillation loss from the task loss. Overall, these variants of knowledge distillation have shown promising results in improving the performance of knowledge distillation.
\end{abstract}

\maketitle

\section{Introduction}
Knowledge distillation~\cite{ref10_kd} is a method of transferring the knowledge from a complex model, called the teacher, to a smaller and simpler model, called the student. In recent years, several variants of knowledge distillation have been proposed, including teaching assistant distillation, curriculum distillation, mask distillation, and decoupling distillation. This literature review summarizes the recent developments in these variants of knowledge distillation and discusses their strengths and limitations.

Knowledge distillation is a method of compressing a complex deep neural network (DNN) into a smaller and faster DNN while preserving its accuracy. The process of knowledge distillation involves training a smaller DNN, called the student, to imitate the predictions of a larger and more complex DNN, called the teacher. The student network is trained to produce similar results as the teacher network, but with fewer parameters and lower computational cost. Knowledge distillation has been widely used for model compression and acceleration, and has shown great promise in various applications\cite{li2021seg,linas2,li2021nas,li2020explicit}, such as computer vision and natural language processing.

In recent years, several variants of knowledge distillation\cite{li2022norm,Dong2023diswot,dong2023rd} have been proposed and explored to improve the performance of knowledge distillation~\cite{li2022self,dong2023progressive}. One of these variants is teaching assistant distillation, which introduces an intermediate model, called the teaching assistant, between the teacher and the student. The teaching assistant is trained to mimic the behavior of the teacher, and the student is trained to imitate the outputs of the teaching assistant. This approach has been shown to provide better performance than traditional knowledge distillation, as it can better capture the knowledge learned by the teacher.

Another variant of knowledge distillation~\cite{li2022SFF,lishadow} is curriculum distillation, which designs the learning process to follow a curriculum, similar to human education. The curriculum is designed to present easy examples first and gradually increase the difficulty of the examples as the student improves. This approach has been shown to provide better performance than traditional knowledge distillation, especially for tasks that require a lot of prior knowledge.

Mask distillation is a variant of knowledge distillation~\cite{li2022tf,wei2022convformer} that focuses on transferring the attention mechanism learned by the teacher to the student. In mask distillation, the teacher is trained to produce a mask that indicates the importance of each input feature for a particular prediction. The student is then trained to imitate the predictions of the teacher while using the mask to weigh the importance of the input features. This approach has been shown to provide better performance than traditional knowledge distillation, as it can better capture the attention mechanism learned by the teacher.

Finally, decoupling distillation is a variant of knowledge distillation~\cite{lichengp} that decouples the distillation loss from the task loss. In decoupling distillation, the student is trained to imitate the outputs of the teacher on a validation set, while being trained on the task loss on the training set. This approach has been shown to provide better performance than traditional knowledge distillation, as it can better balance the trade-off between preserving the knowledge of the teacher and adapting to the task.

In conclusion, knowledge distillation is a widely used method for compressing and accelerating deep neural networks

\section{Teacher Assistant}
When knowledge distillation is performed, increasing the model size of the teacher network instead makes the knowledge distillation worse. The results are as follows: the student network is a 2-layer CNN, while the teacher network is a 4-10-layer CNN, respectively, and it can be seen from the results that continuing to increase the depth of the teacher network does not necessarily improve the performance of knowledge distillation.
 
To further prove that the gap does affect knowledge distillation, the authors conducted a small comparison experiment by fixing the teacher network as a 10-layer CNN and changing the number of layers of the student network, and the results are as follows.
 
Based on the above observations, it is not difficult to think that a medium-sized network (teacher assistant~\cite{takd}, TA) should be added between the large network and the small network, and the teacher network should be used to distill the TA first, and then the student network should be distilled by the TA.
When TA was added, the distillation effect of the student network was significantly better than the direct distillation of the teacher network.
 
(NOKD in the table indicates no KD, BLKD indicates baseline KD, and TAKD indicates KD with the TA method proposed above)
From the analysis of the following figure, we know that the TA distilled from the teacher network (KD-TA) is better than the TA trained directly (FS-TA).
 
The TA results obtained from multiple distillations are as follows, the more multiple stages the better the effect, but the more time and space consuming.
 
In conclusion, the TA approach used, with layers of distillation, where the next layer absorbs the knowledge of the previous layer and passes it to the next layer, is a good solution to the problem that an overly strong teacher network provides some knowledge beyond what the student network can learn, resulting in lower efficiency, and helps us to select a teacher network of the right size.

\section{Inverse probability weighted distillation}
Generally speaking, the knowledge embedded in categorical losses satisfies the assumption of independent homogeneous distribution, but the probability distribution of teachers' predictions is unbalanced across categories, and the presence of transfer gaps for both "strong" and "weak" teachers hinders the transfer of knowledge in the tail categories of teachers' predictions. Transfer is hampered by the presence of transfer gaps. Therefore, the knowledge embedded in the teachers' predictions does not satisfy the assumption of independent homogeneous distribution across categories. The original KD ignores this transfer gap and assigns constant weights to categorization loss and distillation loss. Since teachers' knowledge is unbalanced, constant sample weights for distillation losses would be a bottleneck in knowledge transfer.
Inverse probability weighted distillation (IPWD)~\cite{ickd} is a simple and effective way to supplement underweighted training samples in the machine domain. The propensity score of the machine domain is first estimated by comparing class-aware predictions with context-aware predictions . Then, IPWD uses the inverse probabilities as distillation loss sample weights to strengthen the weights of the underweighted samples. In this way, IPWD generates a pseudo-population of samples to handle unbalanced knowledge.
There is an imbalance problem in teachers' knowledge. CIFAR100 and ImageNet are used as examples, and the prediction probability sums of the teacher models for different categories in the training samples are counted. For example, a sample of a dog with a soft label of , then the corresponding values are accumulated in class and class, respectively. Figure 1 reflects the difference in distribution between ground-truth and teacher predictions, although the teacher model is trained on balanced data (blue dashed line), its prediction distribution is unbalanced across temperatures.
 
Also, the categories will be divided into 4 groups on CIFAR100 by teacher prediction probability and ranking, and knowledge distillation will be performed separately. As shown in Table 1, the KD achieved better performance in all subgroups compared to normal training. However, the first 25 categories brought much higher gains than the last 25 categories (mean 5.14\% vs. 0.85\%). This demonstrates that there is indeed a category imbalance in the knowledge implied by the teachers' predictions.
Thus, the training samples in the human domain are no longer in the machine domain. Simply assuming that the training set is a perfect transfer set may lead to selection bias: samples matching "head" knowledge are over-represented, while samples matching "tail" knowledge are under-represented. This will inhibit the transfer of "tail" knowledge. Therefore, the inverse probability weighting (IPW) technique can be used to overcome the confusion effect caused by the transfer gap.
In short, IPW is used to generate a pseudo-all sample, assigning larger weights to under-represented samples and smaller weights to over-represented samples to achieve de-biased distillation:
 
IPWD proposes an unsupervised way to estimate the propensity score in the machine domain , using the human domain using the CLS-trained classification head as a reference to compare with the output of the KD-trained classifier to determine whether the samples are underrepresented in the machine domain. The approach in this paper trains an additional classification head to calculate , and the propensity score is calculated as follows:
 
	IPWD The final objective function can be expressed as
 
	IPWD achieves the best performance in the vast majority of cases, and IPWD leads the other KD methods by a greater margin in heterogeneous distillation, with the following experimental results:

\section{Early exit of KD}
By comparing the training curves of ImageNet under distillation and without distillation for ResNet18, we find that KD is initially useful, but later becomes worse than scratch, regardless of whether the teacher network chooses ResNet34 or ResNet50. The following two plots directly negate the validity of KD, so is ImageNet a more difficult dataset to perform knowledge Is ImageNet therefore a more difficult dataset for knowledge distillation?
 
	Based on the above observations, the ESKD method is used, and the results are as follows. At this point, KD does work, but the degradation phenomenon still exists as the network depth increases.
 
The training curve of ESKD~\cite{ESKD} is as follows:
 
The teacher is not fully trained, and the performance of the incompletely trained teacher is the same as the small network. The experiment for the fully trained teacher is set to 200 epochs and 60 epochs will learn the rate (60/200), while the incompletely trained teacher follows the strategy of (15/50, 10/35), etc.
The following figure shows the effect of the incompletely trained teacher used for distillation, and it can be seen that both are better than the fully trained teacher.
 
	The effectiveness of ES teacher is also verified with different student and teacher networks in the following figure:
 
	The first method is to end the KD process early, while the second method is to actively reduce the capacity of the teacher network to match the student network, both of which improve on the original KD.

\section{Decoupled knowledge distillation}
Decoupled knowledge distillation (DKD) ~\cite{ref13_rkd} is a new approach to knowledge distillation, which divides the traditional knowledge distillation loss into two parts: target class knowledge distillation (TCKD) and non-target class knowledge distillation (NCKD). The results show that TCKD conveys knowledge related to the difficulty of the training samples, while NCKD is important for Logit distillation.DKD is proposed as a more efficient and flexible method for implementing TCKD and NCKD, achieving comparable or better results than feature-based methods on image classification and target detection tasks.
	Decoupled knowledge distillation (DKD) reformulates knowledge distillation (KD) as a weighted sum of two parts, one related to the target class and the other unrelated to the target class. The reformulation uses binary probabilities and probabilities between non-target classes to separate predictions that are relevant and irrelevant to the target class. These two parts are named Target Class Knowledge Distillation (TCKD) and Non-Target Class Knowledge Distillation (NCKD), respectively. the weights of NCKD are coupled with the probabilities of the target classes. The reformulation stimulates the study of the individual effects of TCKD and NCKD, revealing the limitations of the classical coupling formulation.
	The role of TCKD and NCKD in logit distillation is that TCKD conveys knowledge related to the target class, while NCKD focuses on knowledge between non-target classes. The effectiveness of TCKD becomes apparent when the training data becomes challenging, while NCKD is crucial for logit distillation. However, the loss weight of NCKD is suppressed by the teacher's confident predictions. A study showed that better performance was obtained using NCKD on a good prediction sample, suggesting that the good prediction sample was more knowledgeable than the other samples. By considering TCKD and NCKD independently, decoupled knowledge distillation (DKD) is proposed to address these issues.	
	The classical knowledge distillation (KD) approach has some limitations, i.e., the knowledge transfer between teacher and student models is coupled and cannot be balanced. Two types of knowledge transfer, difficulty-based and non-target-based, are considered critical. However, the latter is inhibited when teachers are confident in their predictions. To address this problem, researchers have proposed a new approach, called decoupled knowledge distillation (DKD), which considers each type of knowledge transfer independently and allows the weights to be adjusted.DKD provides an efficient and flexible method for logarithmic distillation, and Algorithm 1 provides the pseudo-code to implement it.
 
	Experiments are described next for two tasks, image classification and object detection, using datasets such as CIFAR-100, ImageNet and MS-COCO. experiments explore the effectiveness of the DKD method, which improves the performance of knowledge distillation by decoupling certain factors. The results show that DKD can consistently improve the performance of teacher-student pairs with the same and different architectures, and obtain comparable or even better performance than feature-based distillation methods. DKD also improves target detection performance when combined with feature-based distillation methods.
 
Extensions to DKD, a logit distillation method for image classification and target detection tasks, are discussed next. These extensions include evaluating the training efficiency of DKD compared to other state-of-the-art methods, providing a new perspective on why bigger models are not always better teachers, evaluating the transferability of deep features learned by DKD, and showing visualizations that demonstrate the advantages of DKD. Experimental results show that DKD provides significant improvements on a variety of datasets and tasks.

In summary, this new approach to interpreting logit distillation decomposes the classical KD loss into target class knowledge distillation (TCKD) and non-target class knowledge distillation (NCKD). The limitations of the coupled knowledge decomposition formulation are discussed, and decoupled knowledge distillation (DKD) is proposed as a solution.DKD achieves significant improvements on image classification and target detection tasks on CIFAR-100, ImageNet, and MS-COCO datasets. However, DKD cannot outperform the state-of-the-art feature-based methods for the target detection task. This paper provides guidance for tuning in supplements, but further research is needed.

\subsection{Symmetric temperature scaling}
The classical KD method uses the Kullback-Leibler scatter to minimize the difference between the output probabilities of teachers and students. However, recent studies have shown that more accurate teachers do not necessarily teach better, and the reasons for this remain unknown. In the following, a new asymmetric temperature scaling (ATS) method is proposed to improve the clarity of teacher-provided error class probabilities and make large teachers teach better.
There are various approaches to knowledge distillation, which include transferring knowledge from complex teachers to smaller students through a combination of cross-entropy and distillation loss. Some studies have examined the dependence of knowledge transfer on student and teacher architecture and found that larger models are not always better teachers. Other work has focused on understanding the advantages of knowledge distillation from a principles-based perspective. The paper also explains the notation used, such as softmax functions, logarithms, and probabilities. In addition, a particular phenomenon where disadvantaged students are unable to fully imitate good teachers due to a mismatch of abilities is discussed and explained in detail.
Knowledge distillation can be decomposed into three parts, namely, correct guidance, smooth regularization, and category discrimination. These three terms are measured quantitatively using the target category probability, the mean of the error category probability, and the variance of the error category probability. An appropriate temperature scaling method that incorporates the validity of these three terms simultaneously is presented here. Category defensibility is considered to be the basis of KD.
A theoretical analysis of the application of temperature scaling to knowledge distillation is presented below. The softening probability vector used in KD is analyzed, showing that an increase in temperature leads to a more uniform probability distribution, and the concepts of intrinsic variance (IV) and derived variance (DV) are also introduced to analyze the class-discriminable terms. And further explains why larger teachers cannot teach better, showing that the effectiveness of KD is more related to differences between error classes than to all classes.
A new temperature scaling method, the asymmetric temperature scale (ATS), is proposed here to make the distribution of error classes during distillation more discriminable. the ATS applies different temperatures to logits in correct and incorrect classes, allowing for more flexible and differentiated distillation instruction during instruction. This approach is particularly beneficial when teachers are overconfident, as it reduces the size of correct category logits, increases the diversity of error category logits, and ultimately increases the variance and accuracy of the derivations.
Experiments using different datasets and network architectures are described here, including ResNet, WideResNet, ResNeXt, VGG, ShuffleNet, AlexNet, MobileNet, and DSCNN. the training setup mainly follows previous studies with slight modifications based on the dataset. The learning rate decayed every 30 cycles, with a batch size of 128 for CIFAR and 64 for the other datasets.
The researchers investigated the phenomenon that having more accurate teacher models does not necessarily lead to better student model learning in knowledge distillation. They identified three factors of successful teachers:correct instruction, smooth regularization, and class differentiation. They propose a solution called asymmetric temperature scaling (ATS) to enhance the derived variance of large teachers to make their distillation labels more discriminatory when teaching. Experiments have shown that the method is effective.
So far, researchers are studying why large teachers have difficulty teaching knowledge distillation (KD) effectively. They hope their research provides a new perspective on the KD field, but they do not anticipate any negative social impact from their work

\subsection{DIST}
DIST method~\cite{Self-KnowledgeDistillation}, i.e., better knowledge is extracted from stronger teachers. Existing methods perform poorly when the predictions of students and stronger teachers do not match exactly in KL dispersion. In this paper, we propose a correlation-based loss to explicitly capture the intrinsic inter-class relationships of teachers and extend this matching to the intra-class level. The method is simple and practical, and experiments show that it adapts well to various architectures, model sizes, and training strategies, consistently maintaining state-of-the-art performance on image classification, target detection, and semantic segmentation tasks.
The most commonly used approach is to use KL scatter to match probabilistic prediction scores between teachers and students. However, this approach can be challenging when teacher and student models differ significantly in size or training strategy. Here, a new method, DIST, is proposed to extract intra-class relationships using Pearson correlation coefficients instead of KL scatter to further improve performance. This method outperforms the common KD and state-of-the-art KD methods for various tasks such as image classification, target detection, and semantic segmentation.
The concept of knowledge distillation (KD), which involves transferring knowledge from a pre-trained teacher model to a student model by minimizing the difference between prediction scores, is discussed here, highlighting the importance of balancing raw classification loss and KD loss in training students, and also discussing the challenges of using stronger teachers in KD, which may lead to greater differences between teachers and students and make it difficult to achieve exact matching. Therefore, an easy way of matching teacher-student predictions is proposed here to solve this problem.
Here the "Distillation from a Stronger Teacher" (DIST) approach can be used to improve knowledge distillation in deep learning. It proposes a loose matching prediction based on the relative rank between teachers and students rather than the exact probability value, and introduces inter-class and intra-class relationship loss to transfer the relationships of multiple classes and similarities of multiple instances to one class per instance, respectively. The method achieves significant improvements in the baseline settings for image classification tasks such as CIFAR-100 and ImageNet.
Experimental setups and results for image classification using the DIST method are described here, which outperforms previous knowledge distillation (KD) methods to handle the large differences between teacher and student models. The method is also shown to be effective on stronger training strategies and larger models, such as the state-of-the-art swing - transformer. results on the CIFAR-100 dataset show that DIST outperforms feature-based distillation methods. The next describes experiments performed on the MS COCO object detection dataset, using DIST as an additional supervision on the final prediction of classes. The results show that DIST achieves competitive results on the COCO validation set and significantly outperforms vanilla KD. by combining DIST with mimic, state-of-the-art KD methods designed for object detection can be surpassed. The latter experiments on two different neural network models (DeepLabV3 and PSPNet) are used to perform semantic segmentation on the cityscape dataset. The proposed DIST method is applied to the prediction of classification heads using the ResNet-101 backbone teacher of DeepLabV3. The results show that DIST outperforms existing knowledge distillation methods on the semantic segmentation task. Two types of relations are also proposed:interclass and intraclass, and experiments show that both of them outperform vanilla knowledge distillation. The proposed DIST significantly outperformed vanilla knowledge distillation without ground-truth labels when students were trained only on knowledge distillation loss.
In summary, the new knowledge distillation method, DIST, aims to improve knowledge distillation from a stronger teacher's perspective. In this paper, we address the problem of differences between students and teachers and propose relationship-based loss to solve this problem.The DIST method is simple and effective in dealing with strong teachers and outperforms the state-of-the-art KD method in various benchmarking tasks including object detection and semantic segmentation.

\subsection{Curriculum distillation}
The classical KD method, which minimizes the loss of kl dispersion between two predictions obtained from the teacher/student model, has a fixed temperature in the softmax layer. A new approach, called Course Temperature for Knowledge Distillation (CTKD), is proposed here, which improves the performance of knowledge distillation by gradually increasing the learning difficulty of students through dynamic and learnable temperatures. The loss of distillation between the instructor and the student is maximized by learning the temperature in a reverse gradient during the student's training. The method is easy to implement and achieves overall improvement at negligible additional computational cost.
The concepts of two domains, course learning and knowledge distillation, are discussed here. Course learning is a training strategy that organizes learning tasks into incremental difficulty and has applications in several fields such as computer vision and natural language processing. Knowledge distillation is a method for transferring knowledge from a pre-trained teacher model to a smaller student model. Existing approaches have been designed to transfer knowledge in various forms, including logic-based, representation-based, and relational-based approaches. Temperature hyperparameters control the difficulty level of the distillation process, with lower values concentrated on the maximum logarithm and higher values flattening the distribution. Most of the work fixes temperature as a hyperparameter, but recently MKD proposes to learn temperature through meta-learning. The proposed CTKD method is more efficient than the MKD method and can be seamlessly integrated into the existing KD framework.
The technique of knowledge distillation course temperature (CTKD) consists of a learnable temperature module to predict the suitable temperature for distillation and a gradient inversion layer to invert the gradient of the temperature module during back propagation. The technique uses an easy-to-hard curriculum that progressively increases the learning difficulty for students.
A method called adversarial distillation is involved here, which involves optimizing the student model to minimize task-specific losses and distillation losses. Distillation losses are calculated using a dynamic temperature value called tau. The authors suggest using an adversarial approach to learn the optimal temperature value where the temperature module is optimized in the opposite direction of the student to maximize the distillation loss. The optimization process consists of solving two subproblems alternately using stochastic gradient descent and using a nonparametric gradient inversion layer to implement the adversarial process.
The course temperature approach organizes the distillation task from easy to difficult, gradually increasing the magnitude of the loss  relative to the temperature. This is inspired by course learning, and students learn better when tasks are organized in a meaningful order, starting with basic concepts and gradually presenting more advanced ones. A cosine table is used to gradually increase , and a learnable temperature module is introduced to predict temperature values, either globally or by instance. The sub-instance temperature method has a better representation capability and thus better distillation performance.
The researchers evaluated their CTKD technique on various neural networks (VGG, ResNet, Wide ResNet, ShuffleNet, and MobileNet) and compared it with other distillation frameworks, including vanilla KD, PKT, SP, VID, CRD, SRRL, and DKD. using a standard experimental setup. Evaluation was performed on the CIFAR-100, ImageNet-2012 and MS-COCO datasets. The results are reported as the mean (standard deviation) of three experiments. Here the results of the proposed CTKD method are applied to various tasks such as CIFAR-100 classification, ImageNet image classification and MS-COCO object detection. The method is shown to significantly improve the performance of different student networks and its effectiveness is demonstrated by loss curves, feature separability and learning curves. The method can be applied as a plug-in to existing distillation devices and can improve their performance without increasing the computational cost. The results show that CTKD can further improve the detection performance in MS-COCO target detection.
The study used an ablation study on CIFAR-100 on the effectiveness of distillation method hyperparameters and components to assess the effectiveness of course parameters, strategies, and counteracting temperature and course distillation. The results showed that a smooth increase in task difficulty and the use of a cosine course strategy worked best. Learning the temperature parameters in an adversarial manner also improved distillation performance, and the pairing of adversarial temperature with course distillation was more effective than for a single element.
Course temperature for knowledge distillation (CTKD) utilizes a dynamic, learnable temperature to organize knowledge distillation tasks from easy to difficult. The temperature is learned in an adversarial manner during student training to maximize distillation losses. CTKD can be easily integrated into existing knowledge distillation frameworks and leads to general improvements at negligible additional computational cost.

\subsection{Masked Generative Distillation}
The Masked Generative Distillation (MGD) method~\cite{Yue2020MatchingGD} guides the student's feature recovery by masking random pixels and forcing it to generate the teacher's full features through a simple block.MGD is a general feature-based distillation method that can be used for different tasks. It is demonstrated here that MGD achieves excellent improvements on various models for a wide range of datasets such as image classification, target detection, semantic segmentation, and instance segmentation. This paper provides impressive results, such as improving the ImageNet top-1 accuracy of ResNet-18 from 69.90\% to 71.69\%.

	Masked Generative Distillation (MGD) involves students generating the teacher's features using the teacher's masked features instead of mimicking it directly. This approach improves the representation of student features and has been shown to bring considerable improvements for a variety of tasks, including image classification, object detection, semantic segmentation, and instance segmentation. The method is simple to use and has only two hyperparameters. The effectiveness of the method on different datasets has been verified through a large number of experiments.
 
	Different approaches to knowledge distillation in machine learning are discussed here, where knowledge is transferred to a student model using a teacher model. The different approaches include extracting information from intermediate layers, attention transfer and contrast learning, and specific applications of knowledge distillation in object detection and semantic segmentation are also mentioned, where the challenge is to determine where to extract information from because of the imbalance between foreground and background.
	MGD generates the teacher's feature map by using a random mask to overlay the student's feature map and then tries to generate the teacher's feature map using the projector layer with the left side pixels. The proposed MGD distillation loss formulation is designed to make the students generate the teacher's features instead of imitating them.MGD can be easily applied to a variety of tasks and is effective in both classification and intensive prediction tasks.The total loss of MGD is a combination of the original loss and distillation loss.
 
	The next experiments use the Masked Generation Distillation (MGD) feature-based distillation method for various tasks including classification, object detection, semantic segmentation, and instance segmentation. For the classification task, the method is evaluated on the ImageNet dataset and distillation loss is computed on the last feature map of the backbone. The results show that MGD provides a significant improvement in accuracy compared to other feature-based and logarithm-based knowledge distillation methods. The hyperparameters used are  and the model is trained for 100 epochs using the SGD optimizer. The experiments were conducted using Pytorch-based MMClassification and MMRazor. Experiments on the COCO2017 dataset for target detection and instance segmentation are described next. Various distillation methods are compared and the results are evaluated for average precision. The authors used the MMDetection framework with an inheritance strategy to initialize the students with the teacher's parameters. The results show that their proposed method, Masked Generation Distillation (MGD), outperforms other state-of-the-art methods in terms of target detection and instance segmentation. The ResNet-50-based retinal net and SOLO models achieved significant improvements with MGD, improving by 3.6 Boundingbox mAP and 3.1 Mask mAP, respectively, on the COCO dataset.The following describes a study on semantic segmentation in which the researchers evaluated their method using the CityScapes dataset. They trained a teacher model (PspNet-Res101) and two student models (PspNet-Res18 and deeplabp3 - res18) and used distillation to transfer knowledge from the teacher to the student. The results showed that their method outperformed state-of-the-art distillation methods, with both homogeneous and non-homogeneous distillation resulting in significant improvements for students. The researchers used MGD, a feature-based distillation method, and combined it with a logit-based distillation method to further improve the results. The model was evaluated using the mIoU metric, and experiments were conducted using MMSegmentation.
	Researchers are comparing two approaches to extracting knowledge from teacher networks and student networks. One method directly mimics the teacher's feature map, while the other uses Masked Generation Distillation (MGD), which forces students to use their masked features to generate the teacher's full feature map. The researchers found that students achieved better performance and accuracy with MGD even when the teacher was themselves, while the improvement from direct imitation was negligible. The researchers also visualized training loss curves for both methods, showing that the students' feature maps gained stronger representation using MGD. At the time of the study, random channels in the masked global distillation (MGD) image classification method were masked instead of spatial pixels. They found that this method improved the performance of the student model. They also investigated the effect of using different teachers on knowledge distillation and found that stronger teachers with similar architectures were better suited for feature-based distillation, while teachers with high accuracy but different architectures were not as effective. The researchers used a generative block called MGD to recover features using two convolutional layers and one activation layer, ReLU. they explored the effects of different compositions of generative blocks and chose the architecture with two convolutional layers and one activation layer. The method can also be applied to other stages of the model, and refining deeper stages is more beneficial for students. The current study is insensitive to the hyperparameter alpha, but when the lambda is less than 0.5, the students have higher performance with larger ratios, while when the lambda is too large, the semantic information on the left side is too poor to generate a complete feature map.
	In summary, the masked generation distillation (MGD) method allows students to use the teacher's masked features to generate the teacher's features instead of directly imitating the teacher's features. The algorithm has enhanced representational capabilities and can be applied to various tasks such as image classification, target detection, semantic segmentation, and instance segmentation. Numerous experiments including on various models and datasets have demonstrated the simplicity and effectiveness of the method.
 
\section{Simple Knowledge Distillation}
	The use of Simple Knowledge Distillation (SimKD)  allows the compression of powerful teacher models into lightweight student models without sacrificing performance. This is achieved by reusing the discriminative classifier of the teacher model for student inference and training the student encoder by feature alignment with a single loss. A projector was also developed to help match the student encoder with the teacher classifier, making the technique applicable to a variety of teacher and student architectures. Experiments show that with the addition of the projector, the technique achieves state-of-the-art results at the cost of a reduced compression ratio.
	Vanilla KD aligns the logit or class projections of the two models, but the performance gap between the original teacher model and the refined student model is still large. Various methods have been proposed to overcome this problem, but they require detailed knowledge representation and optimized hyperparameters. The SimKD technique is proposed, which trains the student model by feature alignment in the previous layer of the classifier and directly replicates the teacher classifier for student inference. On a standard benchmark dataset, SimKD outperforms all state-of-the-art methods compared with various combinations of teacher-student architectures.
	The concept of knowledge distillation (KD), a technique for compressing knowledge from a powerful teacher model into a smaller student model, is discussed here. Transferred knowledge, usually in the form of soft targets, is thought to capture the relationships between different categories and serve as effective regularization during student training. Feature distillation is a common solution to prevent performance degradation in teacher-to-student compression, which utilizes more information from the middle layer of the teacher model. One of the proposed methods, SimKD, is related to hypothetical transfer learning (HTL) and aims to close the performance gap between teachers and students on the same dataset.
	Here, by analyzing the limitations of the existing method and focusing on its improvement, a new method, logit distillation, is proposed, and the knowledge distillation loss function is re-represented as a weighted sum of two components, target class knowledge distillation (TCKD) and non-target class knowledge distillation (NCKD). The effectiveness of each component is investigated and compared with vanilla knowledge distillation.
	A new technique called simple knowledge distillation (SimKD) is discussed here, which allows for state-of-the-art results without the need for complex knowledge representations or hyperparameters.SimKD involves student inference using a pre-trained teacher classifier, which eliminates the need for labeling information and makes feature alignment the only source of generated gradients. The discriminative information contained in the teacher classifier is considered important here, but has been neglected in the knowledge distillation (KD) literature. It is assumed that there is ability-invariant information in the data that can be easily obtained across different models, while the teacher model contains additional necessary ability-specific information that is difficult to obtain for a simple student model. Empirical evidence is provided here to support this hypothesis and to show that SimKD greatly mitigates the performance degradation in teacher-student compression.

\subsection{Summary}

In this paper, we reviewed recent developments in knowledge distillation\cite{li2020explicit,detkd,Segkd,sskd}, a technique for compressing complex deep neural networks (DNNs) into smaller and faster DNNs while preserving accuracy. We discussed four variants of knowledge distillation, including teaching assistant distillation, curriculum distillation, mask distillation, and decoupling distillation. Teaching assistant distillation introduces an intermediate model between the teacher and the student, while curriculum distillation designs the learning process to follow a curriculum. Mask distillation focuses on transferring the attention mechanism learned by the teacher, and decoupling distillation decouples the distillation loss from the task loss. These variants of knowledge distillation have shown promise in improving the performance of knowledge distillation, making it a valuable tool for model compression and acceleration.

  \bibliographystyle{ACM-Reference-Format}
  \bibliography{sample-base}

\end{document}